\documentclass{sigkddExp}

\usepackage{algorithm2e}

\begin{document}

\title{Sentiment Analysis for Twitter : Going Beyond Tweet Text}

 \numberofauthors{1}
\author{
\alignauthor
Lahari Poddar \hspace{2cm} Kishaloy Halder \hspace{2cm} Xianyan Jia\\
       \affaddr{School of Computing}\\
       \affaddr{National University of Singapore}\\
       \email{\{lahari, kishaloy, jiaxiany\}@comp.nus.edu.sg}
}

\maketitle

\begin{abstract}
Analysing sentiment of tweets is important as it helps to determine the users' opinion. Knowing people's opinion is crucial for several purposes starting from gathering knowledge about customer base, e-governance, campaignings and many more. In this report, we aim to develop a system to detect the sentiment from tweets. We employ several linguistic features along with some other external sources of information to detect the sentiment of a tweet. We show that augmenting the 140 character-long tweet with information harvested from external urls shared in the tweet as well as Social Media features enhances the sentiment prediction accuracy significantly. 

\end{abstract}

\section{Introduction}
Analysing sentiment from text is a well-known NLP problem. Several state-of-the-art tools exist that can achieve this with reasonable accuracy. However most of the existing tools perform well on well-formatted text. In case of tweets, the user generated content is short, noisy, and in many cases ($\sim 30\%$) doesn't follow proper grammatical structure. Additionally, numerous internet slangs, abbreviations, urls, emoticons, and unconventional style of capitalization are found in the tweets. As a result, the accuracy of the state-of-the art NLP tools decreases sharply. In this project, we develop new features to incorporate the styles salient in short, informal user generated contents like tweets. We achieve an F1-accuracy of $\sim 71.3\%$ for predicting the sentiment of tweets in our data-set. We also propose a method to discover new sentiment terms from the tweets.

In section \ref{analysis} we present analysis of the data-set. We describe the data-preprocessing that we have done in section \ref{preprocessing}. In section \ref{framework} we describe how the feature-set was extracted, the classification framework, and also the tuning of the parameters for reasonable accuracy. In section \ref{evaluation} we report the performance of our system. We also report how the different features affect the accuracy of the system. We describe how we harvest new sentiment terms using our framework in section \ref{enhancements}. In this section we also present how we predict strength of sentiment from the tweets. We finally conclude with some possible future directions of work in section \ref{conclusion}.

\section{Data-analysis} \label{analysis}
Tweets are short messages, restricted to 140 characters in length. Due to the nature of this microblogging service (quick and short messages), people use acronyms, make spelling mistakes, use emoticons and other characters that express special meanings. Following is a brief terminology associated with tweets:
    \begin{itemize}
    \item Emoticons: These are facial expressions pictorially represented using punctuation and letters. They express user's mood.
    \item Mention: The “@” symbol is used to refer to other users on the microblog.
    \item Hashtags: Users commonly use hashtags to mark topics. This is primarily done to increase the visibility of their tweets.
    \item Url: Because of the short nature of tweet, people use external link(s) to provide additional information (in support of their tweet).
    \end{itemize}
Our dataset contains tweets about `ObamaCare' in USA collected during march 2010. It is divided into three subsets (train, dev, and test). Some tweets are manually annotated with one of the following classes. \\

\quad \textbf{positive, negative, neutral, unsure,} and \textbf{irrelevant}\\

We ignore the tweets which are annotated \textit{unsure}, or \textit{irrelevant}. We present some preliminary statistics about the training data and test data in Table \ref{table:preliminary}. We observe that there is an imbalance in the  dataset. In training dataset, the ratio of positive tweets to negative ones is almost 1:2. In test set, it is heavily skewed with the ratio being less than 1:3. We handle this data imbalance problem using class prior parameters of the learning algorithm. We discuss this is detail in section \ref{classPrior}.

        \begin{table}
        \centering
        \caption{Preliminary Statistics}
        \label{table:preliminary}
        \begin{tabular}{|l|l|} 
        \hline
          \textbf{type of tokens} & \textbf{count} \\ \hline
          all tokens & 40049 \\ \hline
          noun & 14485 \\ \hline
          adj & 1320 \\ \hline
          adv & 1063 \\ \hline
          verb & 3459 \\ \hline
          strong positive &401 \\ \hline
          strong negative & 359 \\ \hline
          weak positive & 480 \\ \hline
          weak negative & 433 \\ \hline
          capitalized words & 2022\\ \hline
          mention(@) & 462\\ \hline
          hashtag(\#) & 4330\\ \hline
          RT & 573\\ \hline
          positive emoticons & 8\\ \hline
          positive emoticons & 3\\ \hline
        \end{tabular}
        \end{table}

\section{Data pre-processing} \label{preprocessing}
Since tweets are informal in nature, some pre-processing is required. Consider the following tweet.\\
\textbf{\small{``\#Healthcare \#Ins. Cigna denies \#MD prescribed \#tx 2 customers 20\% of the time. - http://bit.ly/5PoQfo \#HCR \#Passit \#ILDems \#p2 PLS RT"}}

It is difficult to understand what is the content of the tweet unless it is normalized. We process all the tweets through the following stages.

\subsection{Normalization}
    Normalization is done as follows: 
    \begin{enumerate} 
    \renewcommand{\labelenumi}{(\theenumi)}
    \item Removing patterns like 'RT', '@user\_name', url.
    \item Tokenizing tweet text using NLTK\cite{bird2006nltk} word tokenizer.
    \item Making use of the stopwords list by NLTK to remove them from the tweet text.
    \item Rectifying informal/misspelled words using normalization dictionary \cite{han-cook-baldwin:2012:EMNLP-CoNLL}. For example, ``foundation" for ``foudation", ``forgot" for ``forgt".
    \item Expanding abbreviations using slang dictionary\footnote{Slang Dictionary - Text Slang \& Internet Slang Words. http://www.noslang.com/dictionary/}. For example, ``btw" is expanded to ``by the way".
    \item Removing emoticons. However we keep the number of positive and negative emoticons in each tweet as feature. We make use of the emoticon dictionary(Table \ref{table:emoticons}) presented in \cite{vashishtfacebook}. 
    \end{enumerate}

        \begin{table}
        \centering
        \caption{Sample from dictionary of emoticons}
        \label{table:emoticons}
        \begin{tabular}{|l|l|l|} \hline
        Emoticons & Classification & Sentiment \\ [0.5ex] 
         \hline\hline
         :-) =) :) :] & Happiness & Positive\\
         \hline
         :-( =( :[ :( & Sadness & Negative\\
         \hline
         ;-) ;) & Wink & Positive\\[1ex] 
        \hline
        \end{tabular}
        \end{table} 
\subsection{Hashtag Segmentation}
We segment a hashtag into meaningful English phrases. The `\#' character is removed from the tweet text. As for example, \textbf{\#\textit{killthebill}} is transformed into \textbf{\textit{kill the bill}}.

In order to achieve this, we use a dictionary of English words. We recursively break the hashtagged phrase into segments and match the segments in the dictionary until we get a complete set of meaningful words. This is important since many users tend to post tweets where the actual message of the tweet is expressed in form of terse hashtagged phrases.
    
\subsection{Processing URLs}
The urls embedded in the tweet are a good source of additional context to the actual short tweet content. Sometimes tweets are too terse to comprehend just from the text content of it alone. However if there is a url embedded in the tweet, that can help us understand the context of it -- perhaps the sentiment expressed as well.\\
In order to leverage this additional source of information, we identify all the urls present in the tweets and crawl the web-pages using AlchemyAPI\footnote{http://www.alchemyapi.com/api}. The API retrieves only the textual body of the article in a web-page. We analyze the article texts later on to get more context for the tweet.
    
\section{Algorithmic Framework} \label{framework}

We employ a supervised learning model using the manually labeled data as training set and a collection of handcrafted features. In this section we describe the features and the classification model used in this task.

    \subsection{Feature Extraction}  

    \begin{table*}
    \centering
    \caption{Features}    
    \label{table:feature}
    \begin{tabular}{|l|l|l|l|}
    \hline 
    Basic   & POS tag & $f_1$    & \# of noun, adj, adv, verb \\ \cline{2-4}                                       
            & Word Polarity  & $f_2$  & \begin{tabular}[c]{@{}l@{}}\# of Strong Positive Words, Strong Negative Words\\ \# of Weak Positive Words, Weak Negative Words\end{tabular} \\ \hline 
    Advanced     & Twitter specific & $f_3$ & Whether the tweet is a retweet or not, contains user mention or not \\ \cline{2-4}   
      & Emoticon & $f_4$ & \# of positiveEmoticons, negativeEmoticons \\ \cline{2-4} 
            & Url  & $f_5$    & \begin{tabular}[c]{@{}l@{}}fraction of positive, negative and neutral sentiment \\sentences in the landing page of the url (if any) \end{tabular}\\ \cline{2-4}
            & Hashtag & $f_6$ & \# of hashtags \\ \cline{2-4}
            & Capitalization & $f_7$ & \# of capitalization word in each tweet \\ \cline{2-4}
             & TF-IDF & $f_8$  &    Stacked predictions from Tf-Idf features \\ \cline{2-4}
             & User & $f_{9}$ &  User id of the user posting the tweet \\ \hline
    \end{tabular}
    \end{table*}

    Table \ref{table:feature} presents the set of features we use in our experiment. We have used some basic features (that are commonly used for text classification task) as well as some advanced ones suitable for this particular domain.
    
    \subsubsection{Basic Features}
    We use two basic features: 
    \begin{enumerate} 
    \renewcommand{\labelenumi}{(\theenumi)}
    \item \textbf{Parts of Speech (POS) tags:} We use the POS tagger of NLTK to tag the tweet texts \cite{bird2006nltk}. We use counts of noun, adjective, adverb, verb words in a tweet as POS features.
    \item \textbf{Prior polarity of the words:} We use a polarity dictionary \cite{wilson2005recognizing} to get the prior polarity of words. The dictionary contains positive, negative and neutral words along with their polarity strength (weak or strong). The polarity of a word is dependent on its POS tag. For example, the word `\textbf{\textit{excuse}}' is negative when used as `noun' or `adjective', but it carries a positive sense when used as a `verb'. We use the tags produced by NLTK postagger while selecting the prior polarity of a word from the dictionary. We also employ stemming (Porter Stemmer implementation from NLTK) while performing the dictionary lookup to increase number of matches. We use the counts of weak positive words, weak negative words, strong positive words and strong negative words in a tweet as features.
    \end{enumerate}

    \subsubsection{Advanced Features}
    We have also explored some advanced features that helps improve detecting sentiment of tweets.
    \begin{enumerate} 
    \renewcommand{\labelenumi}{(\theenumi)}
    \item \textbf{Emoticons:} We use the emoticon dictionary from\cite{vashishtfacebook}, and count the positive and negtive emocicons for each tweet.
    \item \textbf{The sentiment of url:} Since almost all the articles are written in well-formatted english, we analyze the sentiment of the first paragraph of the article using Standford Sentiment Analysis tool\cite{socher2013recursive}. It predicts sentiment for each sentence within the article. We calculate the fraction of sentences that are negative, positive, and neutral and use these three values as features. 
    \item \textbf{Hashtag:} We count the number of hashtags in each tweet.
    \item \textbf{Capitalization:} We assume that capitalization in the tweets has some relationship with the degree of sentiment. We count the number of words with capitalization in the tweets.
    \item \textbf{Retweet:} This is a boolean feature indicating whether the tweet is a retweet or not.
    \item \textbf{User Mention:} A boolean feature indicating whether the tweet contains a user mention.
    \item \textbf{Negation:} Words like `no', `not', `won't' are called negation words since they negate the meaning of the word that is following it. As for example `good' becomes `not good'. We detect all the negation words in the tweets. If a negation word is followed by a polarity word, then we negate the polarity of that word. For example, if `\textit{\textbf{good}}' is preceeded by a `\textit{\textbf{not}}', we change the polarity from `\textit{\textbf{weak positive}}' to `\textit{\textbf{weak negative}}'.
    \item \textbf{Text Feature:} We use tf-idf based text features to predict the sentiment of a tweet. We perform tf-idf based scoring of words in a tweet and the hashtags present in the tweets. We use the tf-idf vectors to train a classifier and predict the sentiment. This is then used as a stacked prediction feature in the final classifier.
    \item \textbf{Target:} We use the target of the tweet as categorical feature for our classifier.
    \item \textbf{User:} On a particular topic one particular user will generally have a single viewpoint (either positive or negative or neutral). If there are multiple posts within a short period of time from a user, then possibly the posts will contain the same sentiment. We use the user id as a categorical feature. On an average there are $3.5$ tweets per user in the dataset, and over $90\%$ users in the train set have expressed a single viewpoint for all their tweets (either positive or negative). Hence we believe this feature should be able to capture a user's viewpoint on the topic.
    \end{enumerate}.

    \subsection{Classifier}
    We experiment with the following set of machine learning classifiers. We train the model with manually labeled data and used the above described features to predict the sentiment. We consider only \textbf{positive}, \textbf{negative} and \textbf{neutral} classes.
    
    \begin{enumerate} 
    \renewcommand{\labelenumi}{(\theenumi)}
    \item \textbf{Multinomial Naive Bayes} : Naive Bayes have been one of the most commonly used classifiers for text classification problems over the years. Naive Bayes classifier makes the assumption that the value of a particular feature is independent of the value of any other feature, given the class variable. This independence assumption makes the classifier both simple and scalable. Bayes classifier assigns a class label $\widehat{y}= C_{k}$ for some k according to the following equation:
    \begin{equation}
    \widehat{y}= \underset{k\in \left \{ 1,...,K \right \}}{argmax}\: \:  p\left ( C_{k} \right ) \prod_{i=1}^{n} p\left ( x_{i} | C_{k}\right )
    \end{equation}

The assumptions on distributions of features define the event model of the Naive Bayes classifier. We use multinomial Naive Bayes classifer, which is suitable for discrete features (like counts and frequencies).

    \item \textbf{Linear SVM} : Support Vector Machines are linear non-probabilistic learning algorithms that given training examples, depending on features, build a model to classify new data points to one of the probable classes. We have used support vector machine with stochastic gradient descent learning where gradient of loss is estimated and model is updated at each sample with decreasing strength.
       \end{enumerate}.
For this task we found Multinomial Naive Bayes performs slightly better than Linear SVM, hence in the evaluation we report accuracy with this classifier.   
    
\subsection{Parameter Tuning}\label{classPrior}
Parameter tuning or hyperparameter optimization is an important step in model selection since it prevents the model from overfitting and optimize the performance of a model on an independent dataset. We perform hyperparameter optimization by using grid search, i.e. an exhaustive searching through a manually specified subset of the hyperparameter space for a learning algorithm. We do grid search and set the `best parameters' by doing cross validation on training set and verified the improvement of accuracy on the validation set. Finally we use the model with best hyperparameters to make predictions on the test set.

\begin{table*}[ht]  
		\centering
		\caption{Experimental results for various features}        
		\label{table:result}        
		\begin{tabular}{|l|l|l|l|l|l|l|l|l|l|l|l|l|}         
		\hline
		&  \multicolumn{3}{l|}{avg/total}&\multicolumn{3}{l|}{Positive}&\multicolumn{3}{l|}{Neutral}&\multicolumn{3}{l|}{Negative}\\ \hline
		Feature combination & Prec & Recall & F1 & Prec & Recall & F1  & Prec & Recall & F1  & Prec & Recall & F1 \\ \hline
		+f1,f2 & 0.57 & 0.45 & 0.47 & 0.32 & 0.43 & 0.37 & 0.25 & \textbf{0.57} & 0.35 & 0.72 & 0.42 & 0.53  \\ \hline
		+f1,f2,f4 & 0.57 & 0.45 & 0.47 & 0.33 & 0.44 & 0.38 & 0.25 & 0.56 & 0.34 & 0.72 & 0.42 & 0.53  \\ \hline
		+f1,f2,f4,f6,f7 & 0.59 & 0.46 & 0.49 & 0.30 & 0.43 & 0.35 & 0.26 & 0.54 & 0.35 & 0.76 & 0.45 & 0.57  \\ \hline
	+f1,f2,f4,f5,f6,f7 & 0.60 & 0.51 & 0.53 & 0.33 & 0.45 & 0.38 & 0.28 & 0.50 & 0.36 & 0.75 & 0.52 & 0.62  \\ \hline
		+f1,f2,f3,f4,f5,f6,f7,f9 & 0.63 & 0.57 & 0.59 & 0.31 & \textbf{0.56} & 0.40 & 0.36 & 0.30 & 0.32 & \textbf{0.79} & 0.64 & 0.71  \\ \hline
		+f1,f2,f3,f4,f5,f6,f7,f8,f9 &\textbf{ 0.69} & \textbf{0.71} & \textbf{0.69} & \textbf{0.51} & 0.33 & \textbf{0.40} & \textbf{0.66} & 0.35 & \textbf{0.45} & 0.75 & \textbf{0.91} & \textbf{0.82}  \\ \hline
		\end{tabular}
\end{table*}

\section{Evaluation and Analysis}\label{evaluation}

Table \ref{table:result} shows the test results when features are added incrementally. We start with our basic model (with only POS tag features and word polarity features) and subsequently add various sets of features. First we add emoticon features, it has not much effect. This is reasonable since only 8 positive emoticons and 3 negative emoticons are detected(Table \ref{table:preliminary}) out of 40049 tokens. So the significance of emoticon can be neglected in this dataset. Then we add hashtag and capitalization features, and obtain an overall gain of 2\% over the basic model. By adding the sentiment features from URL articles, we get overall 6\% improvement over baseline. Further twitter specific features and user features improve the f1 by 12\%. Last, we add TF-IDF feature, and the result improves a lot, and our sentiment classifier reaches the best classification results with an F1-accuracy of $69\%$ as shown in the table. 


Analyzing the results for different classes, we observe that the classifier works best for negative tweets. This can be explained by the number of training tweets for each classes, since proportion of negative tweets were considerably higher in both train and test sets as mentioned in Section \ref{analysis}.

\subsection{Comparison with Stanford Sentiment Analysis Tool}
In this section we compare the performance of our framework with an openly available state-of-the-art sentiment analysis tool. We choose Stanford coreNLP package as the baseline. It uses recursive deep models to do sentiment analysis and achieves good accuracy ($\sim 85\%$) for formal corpora \cite{socher2013recursive}. However for noisy and informal texts like tweets, their performance decreases sharply. We present the performance of Stanford coreNLP tool over the test dataset.

\begin{table}[h]
\centering
\caption{Performance of Stanford CoreNLP tool}
\label{stanfordPerformance}
\begin{tabular}{|l|l|l|l|l|}
\hline
          & precision & recall & F1-score & support \\ \hline
negative  & 0.63      & 0.50   & 0.56     & 493     \\ \hline
neutral   & 0.11      & 0.31   & 0.16     & 127     \\ \hline
positive  & 0.41      & 0.05   & 0.09     & 143     \\ \hline
avg/total & 0.50      & 0.38   & 0.40     & 763     \\ \hline
\end{tabular}
\end{table}

Comparing table \ref{stanfordPerformance} with table \ref{table:result} we observe that our framework outperforms stanford coreNLP by a significant margin ($\sim 20\%$). This owes to the fact that stanford coreNLP is not able to handle text with lot of noise, lack of formality, and slangs/abbreviations. This proves the effectiveness of our framework.
\newpage
\section{Enhancements}\label{enhancements}
Apart from sentiment prediction, we also present some extensions to our system.

\subsection{Harvest New Sentiment Terms}
We have used a static dictionary to get prior polarity of a word, which helps detect the overall sentiment of a sentence. However the usage of words varies depending on conversation medium (e.g. : informal social media, blogs, news media), context and topic. For instance, the word \textit{`simple'} is generally used in \textbf{positive} sense, but consider its use while describing the storyline of a movie. In this context, a \textit{`simple storyline'} will probably hint at a negative sentiment. For a dynamic media like Twitter, where the topic mix and word mix change often, having a static dictionary of words with fixed polarity will not suffice. To get temporal and topic-specific sentiment terms, we make use of the tweets classified by our classifier.

We consider the words that appear in the positive, neutral and negative tweets. A word that very frequently occurs in tweets with positive (negative) sentiment and hardly occurs with tweets with a negative (positive) sentiment, will probably have a positive (negative) orientation for that particular topic. To implement this hypothesis, we first count the word frequency in each tweet collection. Then for each collection, we select top $k\%$ most frequent words and deduct from top $n\%$ words from other two collections. For example, in \textbf{Algorithm} \ref{alg:newsen}, if we want to get new negative words, we find the words in top $10\%$ from negative collection. And we compare the words that appear in top $60\%$ of the other two, remove words that co-appear. Part of the new negative terms we find are shown in \textbf{Table \ref{table:newNegative}}. We use same procedure to find new positive and neutral words.


\begin{algorithm}
\caption{Harvest New Negative Words Algorithm}
\label{alg:newsen}
 \KwData{negativeCol, positiveCol, neutralCol}
 \KwResult{new negative words from data collection }
 
 $NEG_{top} \leftarrow NEG[0:<threshold1>*L_{NEG}]$\;
 $POS_{top} \leftarrow POS[0:<threshold2>*L_{POS}]$\;
 $NEU_{top} \leftarrow NUE[0:<threshold2>*L_{NUE}]$\;
 \For{$word \in NEG_{top}$}{
  \eIf{$word \notin \{POS_{top} \cup NEU_{top}\}$}{
   $NewNagative \leftarrow NewNegative \cup \{word\}$\;
   }{
   drop word
  }
 }
 
\end{algorithm}

\begin{table}
        \centering
        \caption{Part of new negative terms}
        \label{table:newNegative}
        \begin{tabular}{|l|l|} 
        \hline
        term & frequency \\ \hline
        tweetcongress  &   19 \\ \hline
        lie & 15\\ \hline
        txgop  &  11\\ \hline
        taxes  &  11\\ \hline
        tpp&  9\\ \hline
        deem  &   9\\ \hline
        gov't &   8\\ \hline
        vulnerable&   8\\ \hline
        sad & 7\\ \hline
        stupid &  7\\ \hline
        cancer &  7\\ \hline
        unconstitutional  &   7\\ \hline
        \end{tabular}
\end{table}

\subsection{Predicting Strength of Sentiment}
Apart from predicting the sentiment class of tweets we are also interested in predicting the strength or intensity of the sentiment associated. Consider the following tweets.
\begin{itemize}
\item t1: \textbf{\small{`GO TO YOUR US REPS OFFICE ON SATURDAY AND SAY VOTE NO! ON \#HCR \#Obama \#cnn \#killthebill \#p2 \#msnbc \#foxnews \#congress \#tcot'}}
\item t2: \textbf{\small{`Thankfully the Democrat Party isn't too big to fail. \#tcot \#hcr'}}
\end{itemize}

Although both the tweets have negative sentiment towards `ObamaCare', the intensity in both are not the same. The first tweet (t1) is quite aggressive whereas the other one (t2) is not that much. Here we propose a technique to predict the strength of sentiment.

We consider few features from the tweet in order to do this. If our classifier predicts the sentiment to be neutral we say that the strength of sentiment is 0. However if it is not i.e., if it is either positive or negative, we increase strength of sentiment for each of the following features of the tweet.

\begin{enumerate}
\item Number of capitalized words.
\item Number of strong positive words.
\item Number of strong negative words.
\item Number of weak positive words.
\item Number of weak negative words.
\end{enumerate}

Each of these features contributes to the strength score of a tweet. Once calculated, we normalize the score within [0-5]. Finally we assign sentiment polarity depending on the overall sentiment of the tweet. As for example, if a tweet has score of 3 and the overall predicted sentiment is negative then we give it a score of `-3'. It denotes that the tweet is moderately negative. Having said that, strength of sentiment is highly subjective. A tweet can appear to be very much aggressive to some person whereas the same may appear to not to be that aggressive to some other person. 


\section{Conclusion}\label{conclusion}

In this report we have presented a sentiment analysis tool for Twitter posts. We have discussed the characteristics of Twitter that make existing sentiment analyzers perform poorly. The model proposed in this report has addressed the challenges by using normalization methods and features specific to this media. We show that using external knowledge outside the tweet text (from landing pages of URLs) and  user features can significantly improve performance. We have presented experimental results and comparison with state-of-the-art tools.

We have presented two enhanced functionalities, i.e. discovering new sentiment terms and predicting strength of the sentiment. Due to the absence of labelled data we couldn't discuss the accuracies of these two enhancements. In the future, we plan to use these as feedback mechanism to classify new tweets.

%
\bibliographystyle{unsrt}  
\bibliography{reference}
%
%

\end{document}